\documentclass[a4paper]{svproc}
\usepackage{url}

\usepackage{graphicx}
\usepackage{amsmath}
\usepackage{amssymb}
\usepackage{subcaption}
\usepackage{siunitx}
\usepackage{booktabs}
\usepackage{multirow}
\usepackage{algorithm}
\usepackage{algpseudocode}
\usepackage{tikz}
\usetikzlibrary{positioning, shapes.geometric, arrows.meta, calc, fit, backgrounds}

\begin{document}
\mainmatter

\title{Obstacle-Aware Autonomous Coverage and Navigation for Outdoor Robots}

\titlerunning{Obstacle-Aware Autonomous Coverage and Navigation for Outdoor Robots}

\author{
Leonardo Gargani \and
Matteo Frosi \and
Matteo Matteucci
}

\authorrunning{Leonardo Gargani et al.}

\institute{
Department of Electronics, Information and Bioengineering (DEIB),\\
Politecnico di Milano, Italy\\
\email{leonardo.gargani@polimi.it}
}

\maketitle

\begin{figure}[t]
    \centering
    \begin{subfigure}[b]{0.5\linewidth}\centering
        \includegraphics[height=4.0cm]{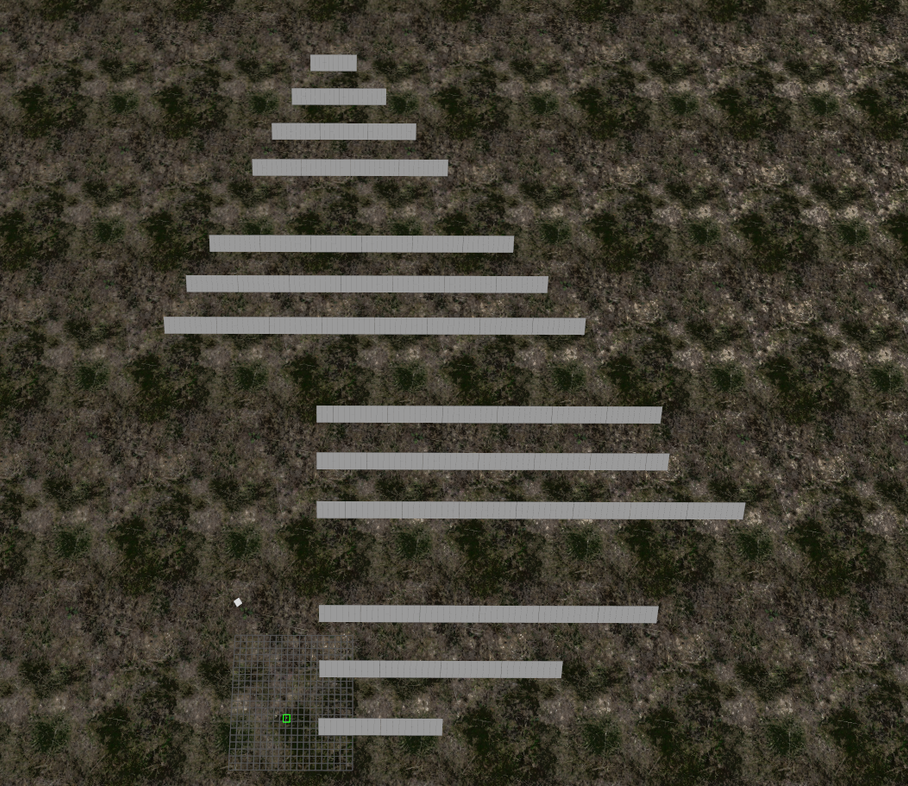}
        \label{fig:front-a}
    \end{subfigure}\hfill
    \begin{subfigure}[b]{0.5\linewidth}\centering
        \includegraphics[height=4.0cm]{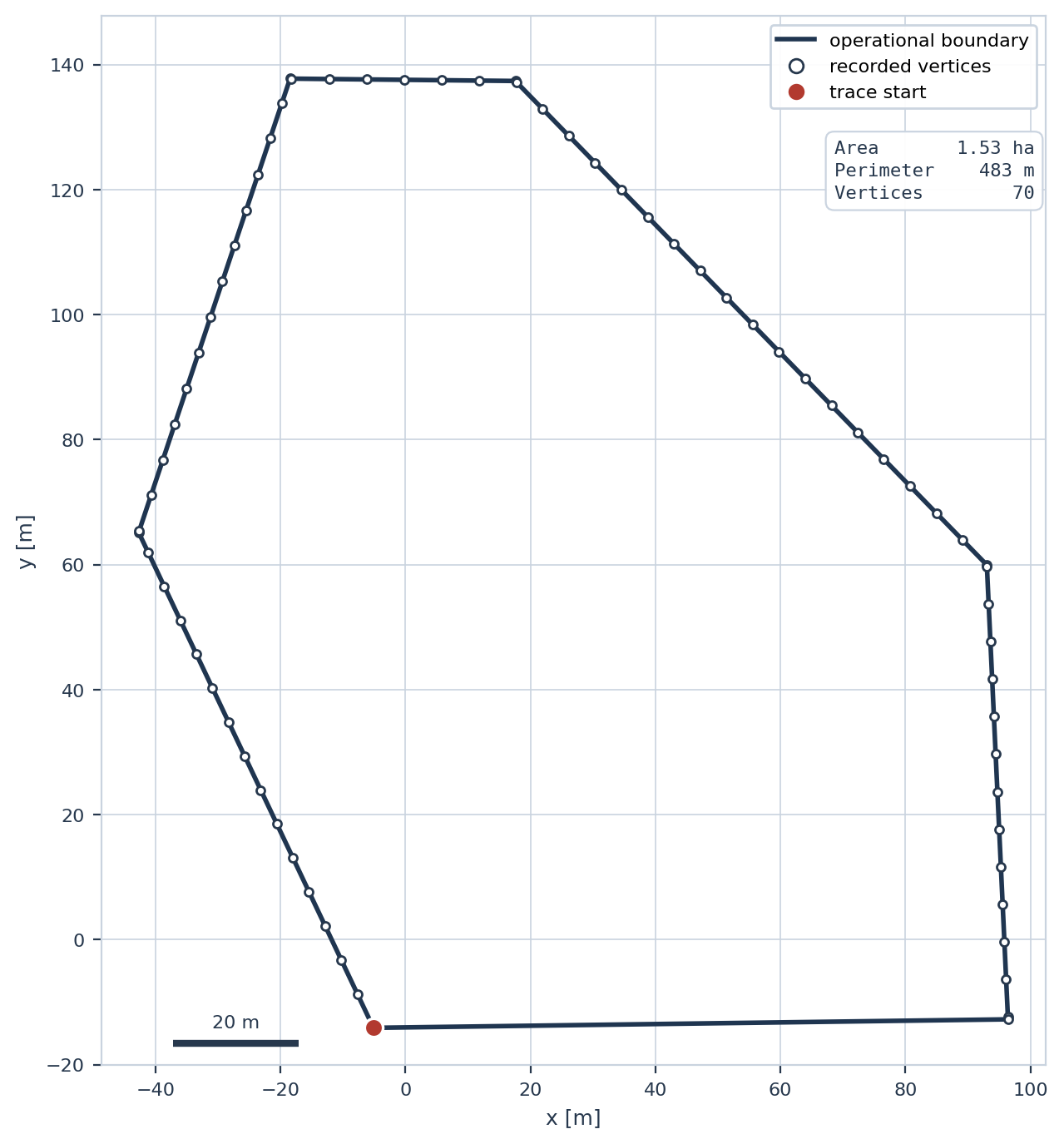}
        \label{fig:front-b}
    \end{subfigure}\\[0.5cm]
    \begin{subfigure}[b]{0.5\linewidth}\centering
        \includegraphics[height=4.0cm]{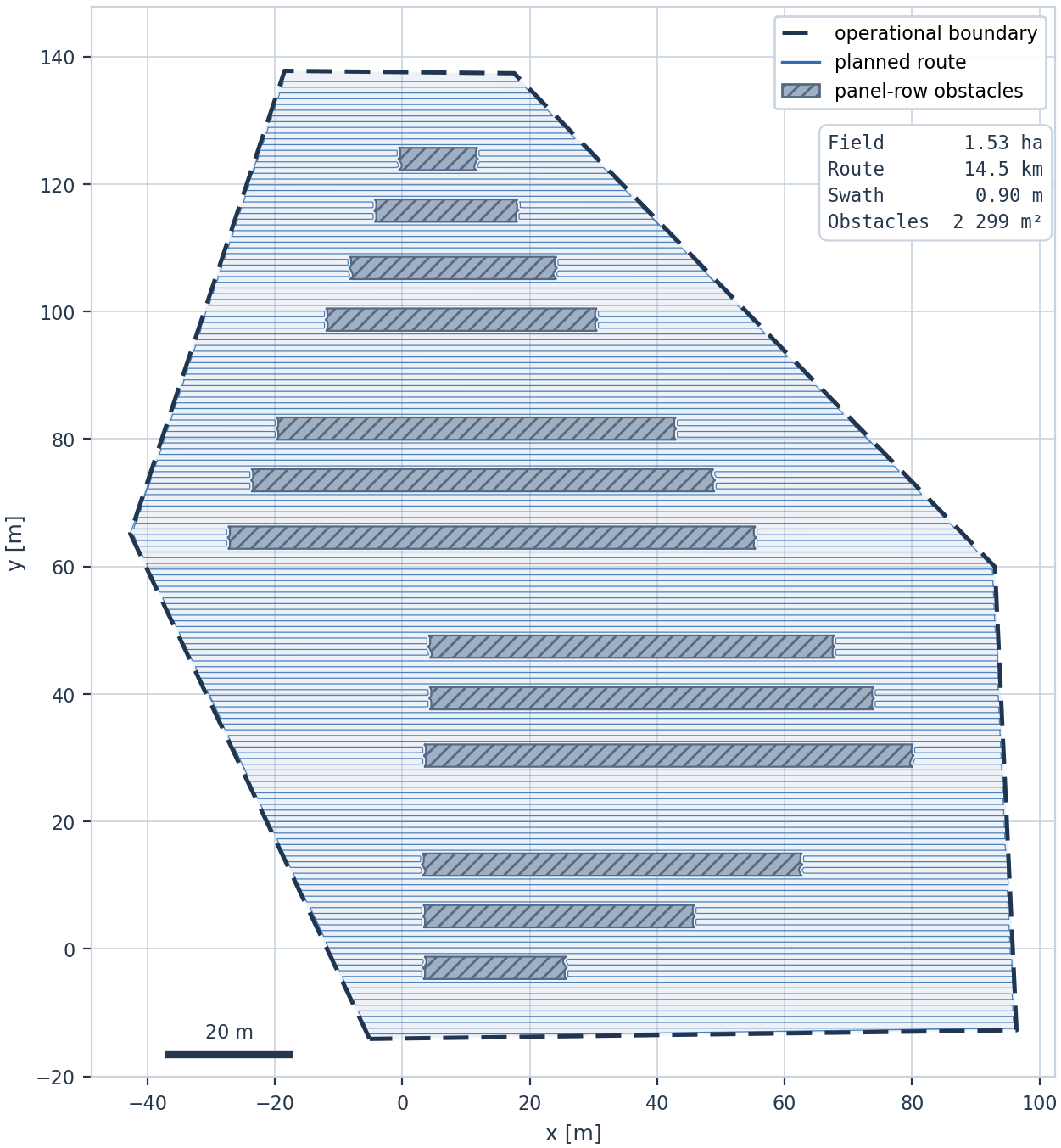}
        \label{fig:front-c}
    \end{subfigure}\hfill
    \begin{subfigure}[b]{0.5\linewidth}\centering
        \includegraphics[height=4.0cm]{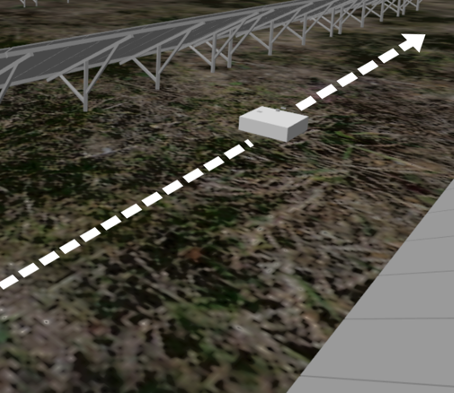}
        \label{fig:front-d}
    \end{subfigure}
    \caption{Stages of the proposed framework, shown from left to right and top to bottom: environment and working-area definition, perimeter recording and admissible-region construction, full coverage computation, and autonomous mission execution.}
    \label{figure:front_page}
\end{figure}

\begin{abstract}

Long-duration outdoor coverage with autonomous platforms remains challenging beyond classical planning: deployments face localization drift in open spaces, obstacles in cluttered sites, controller feasibility in turn-heavy maneuvers, and persistent autonomy with energy management. We propose a unified ROS~2 architecture for single-robot outdoor coverage: a dual-antenna RTK-GNSS fused in an EKF keeps both position and heading accurate across long missions; three controller-aware refinements extend a mature coverage planner; a Nav2-based behavior-tree mission executive coordinates multi-goal execution, layered recovery, cost-aware goal management, and autonomous docking for return-to-charge. In real-world trials across five outdoor areas with varying geometries and obstacle densities, the robot completed every coverage route, sweeping \SIrange[quantity-product={}]{93.1}{96.1}{\percent} of the planned coverage area.


\end{abstract}

\section{Introduction}

Industrial outdoor sites such as solar farms, logistics yards, and large managed green areas increasingly deploy autonomous platforms for repetitive maintenance (mowing, sweeping, inspection, soil conditioning). Coverage strategies are well studied in controlled settings, but reliable field operation remains hard under changing conditions, requiring stable localization in open spaces, safe interaction with known and unknown obstacles, long-duration execution, smooth return-to-charge without operator intervention, and other mission-level capabilities.

These constraints expose three closely related gaps in many existing approaches. First, low-speed motion with frequent turns makes heading hard to estimate, especially during intermittent GNSS degradation. Second, geometrically optimal coverage paths may still be fragile in execution when waypoint distribution and turn resolution do not match controller feasibility. Third, long-duration operation is inherently system-level: planning and control alone need structured supervision, recovery policies, and safe interruption for energy-aware mission management. Modern navigation frameworks such as Nav2 provide behavior-tree-based orchestration~\cite{macenski2020marathon}, yet integrating robust localization, planning refinements, robust execution, and persistent operation into a cohesive, field-deployable architecture remains an open systems integration problem.

We propose a unified ROS~2 architecture for single-robot outdoor coverage under a mission-level supervisor (Fig.~\ref{figure:front_page}). We target the gap between a geometrically valid plan and reliable field execution: we combine perception, robust outdoor localization, controller-aware planning refinements, and a persistent behavior-tree executive, and validate the full system in five real outdoor areas of varying geometry, establishing feasibility. Our contributions follow.

\medskip

\begin{itemize}
\item \textbf{An outdoor localization stack} for long coverage missions: a dual-antenna RTK-GNSS and wheel-odometry fusion in a custom 12-state EKF with per-sensor Mahalanobis gating and a fallback inferring heading from the direction of travel, designed to keep the global pose reliable through tight turns and intermittent GNSS, the regime where wheel odometry alone drifts.
\item \textbf{Controller-aware coverage refinements} layered on Fields2Cover and formalized against the native planner behavior: collinearity-aware resampling that adapts waypoint density to the controller's tracking needs, obstacle-aware connectors that delegate inter-swath transitions to the reactive layer to preserve clearance, and layout-aware obstacle clustering that merges cluttered layouts into clean obstacle polygons before planning.
\item \textbf{A behavior-tree mission executive} integrated with Nav2 that converts an offline coverage route into long-horizon outdoor autonomy in real field deployments: batched multi-goal execution, a layered recovery policy, cost-aware goal management under interference from unknown obstacles, and marker-guided autonomous docking with energy-triggered return-to-charge.
\item \textbf{An end-to-end evaluation} across five real outdoor areas spanning convex and non-convex geometries under realistic conditions, with and without known obstacles, reporting mission completion, run duration, and a coverage-efficiency metric that isolates execution quality from the plan itself.
\end{itemize}

\section{Related Works}
\label{sec:related}

\paragraph{Coverage path planning and field pipelines.} Coverage path planning classically guarantees complete coverage by geometric decomposition, such as boustrophedon cellular decomposition~\cite{choset2000coverage} and related grid and spanning-tree formulations~\cite{gabriely2001spanning}. Since computing a length-optimal complete coverage path is generally costly, practical planners typically decompose the area into cells covered by parallel swaths~\cite{galceran2013survey}. This swath-and-route strategy remains widely used in field deployments, consolidated into reusable pipelines: Fields2Cover (F2C)~\cite{mier2023fields2cover} packages headland and swath generation, routing, and path synthesis in a modular open-source library. Fields2Benchmark~\cite{mier2025fields2benchmark} standardizes evaluation on a large dataset of real fields of varied geometry, while reviews organize coverage into plan generation, swath traversal ordering, and smooth drivable-path planning~\cite{hoffmann2024optimal}.

\paragraph{Coverage transitions and execution robustness.} Headland and inter-swath transitions are an active subproblem, addressed by continuous-curvature headland coverage with clothoid turns at field corners~\cite{mier2025continuous} and by optimization-based motion planning for single headland turns in constrained fields with obstacles~\cite{peng2024optimization}. For execution-time robustness, anytime replanning repairs only the path sections affected by initially unknown obstacles, within a bounded time budget~\cite{ramesh2024anytime}. These pipelines leave open how reliably a real robot executes the plan, keeping clearance from known obstacles and handling unknown ones.

\paragraph{Outdoor localization and heading estimation.} Outdoor coverage execution needs a stable global pose, yet the robot often moves slowly and turns frequently, making heading difficult to estimate from wheel odometry alone and prone to drift over long missions in the field. In the Robot Operating System (ROS), heterogeneous sensor data are often fused by a configurable extended Kalman filter~\cite{moore2015generalized}. Separately, dual-antenna RTK-GNSS can supply the filter with a motion-independent yaw measurement to improve attitude estimation~\cite{ding2025attitude}.

\paragraph{Execution, behavior trees, and persistent autonomy.} Reliable field operation also rests on the execution layer. Nav2~\cite{macenski2020marathon,macenski2023desks} is a widely used ROS~2 navigation substrate with layered costmaps, controllers, recovery behaviors, and behavior-tree (BT) orchestration, favored over finite-state machines for modular, reactive execution~\cite{colledanchise2018behavior}. Long-term indoor deployments demonstrated persistent autonomy with charging and recovery~\cite{hawes2017strands}; recent work formalizes fleet-level charging schedules for long-duration autonomy~\cite{kumar2025persistent}. The closest integration, \texttt{opennav\_coverage}~\cite{opennav_coverage}, exposes F2C as a Nav2 task server with BT nodes, without localization, plan refinement for closed-loop tracking, goal-level unknown-obstacle handling, docking, or reported evaluation on a real outdoor deployment.

\bigskip
\noindent
Three choices distinguish our stack. The execution layer follows a deterministic route, handling unknown obstacles by temporarily deferring blocked goals to the reactive layer rather than modifying the offline plan~\cite{ramesh2024anytime}. Instead of the configurable ROS filter~\cite{moore2015generalized}, the localization stack uses a custom EKF with gated dual-antenna yaw and a direction-of-travel fallback for turn-heavy coverage. Finally, since the differential-drive platform can rotate in place, the planning refinements target waypoint density and inter-swath connector clearance rather than the continuous-curvature headland turns required by car-like platforms.

\section{System Description and Architecture}
\label{sec:system}

We formalize the coverage task and admissible workspace, then present the offline-to-online architecture, hardware platform, and localization stack that maintains accurate global pose over long, turn-heavy missions in open fields.

\subsection{Problem Formulation}
\label{sec:problem}

We consider a single differential-drive robot operating on predominantly planar outdoor terrain. The workspace is described by an outer boundary polygon $\mathcal{B}\subset\mathbb{R}^2$ and a set of $m$ \emph{known} obstacles $\mathcal{O}=\{O_1,\dots,O_m\}$, each $O_i\subset\mathcal{B}$ a simple polygon mapped a priori during preparation. Obstacles appearing only at runtime, whether fixed objects absent from the map or moving agents such as pedestrians, are \emph{unknown} and handled reactively (Sect.~\ref{subsec:autonomous_navigation}). The two classes differ in what is available at offline planning time, not by whether they move.

The robot has a \emph{working width} $w$, the ground width that its maintenance tool covers in a single pass, and a footprint radius $\rho$. Adjacent passes are laid out at a \emph{swath spacing} $d_s\le w$, so they overlap by $w-d_s$. Its planar pose is $\mathbf{x}=(x,y,\psi)\in SE(2)$, a position and a heading. Being differential-drive, it can turn in place, so its minimum turning radius $r_{\min}$ is effectively zero and the plan need not respect a tight curvature bound at swath transitions.

The robot must keep a safety margin $\delta\ge\rho$ from obstacles, large enough to cover the footprint radius $\rho$ plus extra clearance. We grow each obstacle into $O_i^{\delta}=\{\mathbf{q}\in\mathbb{R}^2:\mathrm{dist}(\mathbf{q},O_i)\le\delta\}$, the set of points within $\delta$ of $O_i$, and subtract grown obstacles from $\mathrm{int}(\mathcal{B})$, the interior of the boundary. What remains is the \emph{admissible region}, the free space in which the plan must remain:
\begin{equation}
  \mathcal{A} = \mathrm{int}(\mathcal{B}) \,\setminus\, \bigcup_{i=1}^{m} O_i^{\delta}.
  \label{eq:admissible}
\end{equation}

A coverage plan is the path the robot is expected to follow. We represent it as a curve $\gamma$ parameterized by arc length $s\in[0,L]$, where $s$ is the distance from the start and $L$ is the total path length. For each $s$, the curve gives the robot pose $\gamma(s)=(x,y,\psi)\in SE(2)$. As the robot moves along it, the tool sweeps a ground region during field execution. If $D_w(\mathbf{x})$ denotes the width-$w$ working band centered at the robot and orthogonal to its heading, this region is
\begin{equation}
  S(\gamma) = \bigcup_{s\in[0,L]} D_w\!\big(\gamma(s)\big).
  \label{eq:swept}
\end{equation}

We judge a plan by how completely the robot sweeps the workspace during execution. Writing $S(\gamma)\cap\mathcal{A}$ for the swept area inside the admissible region, and distinguishing the planned route $\gamma_{\mathrm{plan}}$ from the executed trajectory $\gamma_{\mathrm{exec}}$, the \emph{coverage efficiency} is the fraction of planned coverage actually realized
\begin{equation}
\eta_c =
\frac{
\left|S(\gamma_{\mathrm{exec}})\cap S(\gamma_{\mathrm{plan}})\cap\mathcal{A}\right|
}{
\left|S(\gamma_{\mathrm{plan}})\cap\mathcal{A}\right|
}
\in[0,1],
\label{eq:etac}
\end{equation}
where $|\cdot|$ denotes area (in \si{\square\metre}) and both sweeps use the same working width $w$, so $\eta_c$ isolates execution quality from the plan in the reported trials.

We therefore inherit Fields2Cover's cell-decomposition-and-swaths strategy (Sect.~\ref{sec:related}) and, rather than re-deriving an optimal plan, target the gap between a geometrically valid route and its field execution (Sect.~\ref{subsec:full_coverage}). If every swath is executed at full width $w$, the planned route preserves the completeness of the underlying decomposition. The empirical gap to $\eta_c=1$ (Sect.~\ref{sec:eval}) is due to execution effects such as turning, overlap, and discrete waypoint following.

\subsection{Architecture and Platform}
\label{sec:overview}

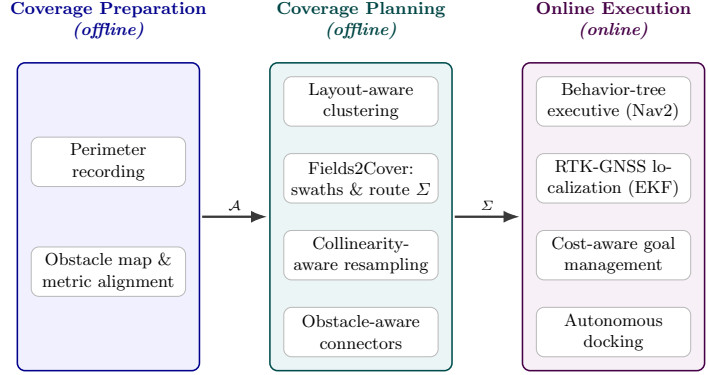
\begin{figure}[t]
    \centering
    \resizebox{0.76\textwidth}{!}{%
    \begin{tikzpicture}[
      font=\footnotesize, >=Latex, node distance=3mm,
      blk/.style={draw=gray!55, rounded corners, fill=white, align=center, minimum height=9mm, inner sep=3pt, text width=26mm},
      fitghost/.style={align=center, minimum height=9mm, inner sep=3pt, text width=26mm},
      modprep/.style={draw=blue!55!black, rounded corners, thick, inner sep=7pt, fill=blue!6},
      modplan/.style={draw=teal!60!black, rounded corners, thick, inner sep=7pt, fill=teal!8},
      modnav/.style={draw=violet!55!black, rounded corners, thick, inner sep=7pt, fill=violet!6},
      lbl/.style={font=\footnotesize\bfseries, align=center},
      ar/.style={->, very thick, draw=gray!45!black, shorten >=1pt, shorten <=1pt}
    ]
    \node[fitghost] (prep_top) at (0,0) {\phantom{Obstacle map \& metric alignment}};
    \node[fitghost] (prep_bot) at (0,-4.2) {\phantom{Obstacle map \& metric alignment}};
    \node[fitghost] (plan_top) at (4.6,0) {\phantom{Collinearity-aware resampling}};
    \node[fitghost] (plan_bot) at (4.6,-4.2) {\phantom{Collinearity-aware resampling}};
    \node[blk] (perim) at (0,-1.1) {Perimeter recording};
    \node[blk] (omap)  at (0,-3.1) {Obstacle map \& metric alignment};
    \node[blk] (clust)  at (4.6,0)     {Layout-aware clustering};
    \node[blk] (f2c)    at (4.6,-1.4)  {Fields2Cover: swaths \& route $\Sigma$};
    \node[blk] (resamp) at (4.6,-2.8)  {Collinearity-aware resampling};
    \node[blk] (conn)   at (4.6,-4.2)  {Obstacle-aware connectors};
    \node[blk] (bt)   at (9.2,0)    {Behavior-tree executive (Nav2)};
    \node[blk] (loc)  at (9.2,-1.4) {RTK-GNSS localization (EKF)};
    \node[blk] (gm)   at (9.2,-2.8) {Cost-aware goal management};
    \node[blk] (dock) at (9.2,-4.2) {Autonomous docking};
    \begin{scope}[on background layer]
      \node[modprep, fit=(prep_top)(prep_bot)(perim)(omap)] (prep) {};
      \node[modplan, fit=(plan_top)(plan_bot)(clust)(f2c)(resamp)(conn)] (plan) {};
      \node[modnav, fit=(bt)(loc)(gm)(dock)] (nav) {};
    \end{scope}
    \node[lbl, text=blue!55!black, above=3mm of prep] {Coverage Preparation\\\emph{(offline)}};
    \node[lbl, text=teal!55!black, above=3mm of plan] {Coverage Planning\\\emph{(offline)}};
    \node[lbl, text=violet!55!black, above=3mm of nav] {Online Execution\\\emph{(online)}};
    \draw[ar] (prep.east) -- node[above, font=\scriptsize]{$\mathcal{A}$} (prep.east -| plan.west);
    \draw[ar] (plan.east) -- node[above, font=\scriptsize]{$\Sigma$} (plan.east -| nav.west);
    \end{tikzpicture}%
    }
    \caption{Architecture of the proposed coverage framework. Offline preparation produces the admissible region $\mathcal{A}$; offline planning turns it into a coverage route $\Sigma$; online execution runs the route under a behavior-tree executive throughout the mission.}
    \label{figure:full_pipe}
\end{figure}

Fig.~\ref{figure:full_pipe} summarizes the offline-to-online framework. Offline, it maps a site geometry to the admissible region $\mathcal{A}$ of Eq.~\eqref{eq:admissible}, clusters the known obstacles, and computes a coverage route $\Sigma$ with Fields2Cover and our refinements (Sect.~\ref{subsec:full_coverage}). Online, a behavior-tree supervisor executes $\Sigma$, with the localization filter maintaining the robot pose and the navigation stack reactively handling unknown obstacles (Sect.~\ref{subsec:autonomous_navigation}). The same stack runs in simulation and on the real robot; the route is computed offline once per site and reused across repeated runs.

Experiments used an outdoor differential-drive robot for large-area maintenance, with wheel encoders, a dual-antenna Unicore UM982 RTK-GNSS with a fixed base station for georeferenced position and heading, a Livox Mid-360 3D LiDAR for obstacle sensing, and a rear camera for docking. The full autonomy stack runs onboard, without cloud services or remote computation.

The deployed coverage planner uses boustrophedon swath ordering at working width $w=\SI{1.0}{\metre}$ and swath spacing $d_s=\SI{0.9}{\metre}$, so adjacent passes overlap \SI{0.1}{\metre} by design. The navigation stack runs a Regulated Pure Pursuit controller~\cite{macenski2023rpp} (nominal speed \SI{0.4}{\metre\per\second}), a grid-based global planner, and a two-stage smoother (curvature-minimizing then Savitzky-Golay) over inflated local and global costmaps~\cite{macenski2023desks}, fed by the localization stack's \SI{20}{\hertz} 12-state EKF.

\subsection{Localization}
\label{subsec:localization}

Accurate heading, a precondition for coverage, is particularly difficult on this platform: heading from wheel motion alone tends to drift in tight turns, accumulating over long missions, eventually degrading the global pose estimate. We fuse wheel odometry with dual-antenna RTK-GNSS giving georeferenced positions and, when available, a motion-independent heading estimate.

A custom EKF performs the fusion; its 12-dimensional state comprises position, orientation, linear velocity, and angular velocity for online estimation,
\begin{equation}
\boldsymbol{\xi}
=
\left[
\begin{array}{cccccccccccc}
x & y & z &
\phi & \theta & \psi &
v_x & v_y & v_z &
\omega_x & \omega_y & \omega_z
\end{array}
\right]^{\top},
\label{eq:ekf_state}
\end{equation}
where $\mathbf{p}=(x,y,z)$ is the robot base position, $(\phi,\theta,\psi)$ roll, pitch, and yaw, $\mathbf{v}=(v_x,v_y,v_z)$ the body-frame linear velocities, and $(\omega_x,\omega_y,\omega_z)$ the angular velocities. On the flat sites considered here the GNSS elevation is unused, so the measurement models are planar and direct: GNSS position corrects $(x,y)$, dual-antenna yaw the heading $\psi$, and wheel odometry, used in differential mode, corrects the forward and yaw velocities. The elevation state is therefore never corrected, which does not affect the planar pose used for coverage. Positions are georeferenced through the fixed base station into the frame shared with the coverage plan, while velocities are expressed in the body frame.

Prediction between measurements uses a constant-velocity model over each filter prediction interval. With $\boldsymbol{\eta}=[\phi,\theta,\psi]^\top$ and $\boldsymbol{\omega}=[\omega_x,\omega_y,\omega_z]^\top$ the orientation and angular-velocity blocks, the prediction reads compactly as
\begin{align}
\mathbf{p}_{k+1}
&=
\mathbf{p}_{k}
+
\mathbf{C}(\phi_k,\theta_k,\psi_k)
\mathbf{v}_{k}\Delta t,
\label{eq:ekf_position_prediction}
\\
\boldsymbol{\eta}_{k+1}
&=
\boldsymbol{\eta}_{k}
+
\mathbf{T}(\phi_k,\theta_k)
\boldsymbol{\omega}_{k}\Delta t,
\label{eq:ekf_orientation_prediction}
\end{align}
where $\mathbf{C}$ maps body-frame velocity into the global frame and $\mathbf{T}$ maps body angular velocity to Euler-angle rates. Covariance is propagated with the Jacobian $\mathbf{F}_k$ of this nonlinear transition at each prediction step to track uncertainty,
\begin{equation}
\mathbf{P}_{k+1}^{-}
=
\mathbf{F}_k
\mathbf{P}_{k}^{+}
\mathbf{F}_k^{\top}
+
\Delta t\,\mathbf{Q},
\label{eq:ekf_cov_prediction}
\end{equation}
with the block-diagonal process noise of Table~\ref{tab:loc_params} used in all experiments.

Innovation gating protects every sensor update during operation. For a measurement $\mathbf{z}_k$ with model $\mathbf{h}(\boldsymbol{\xi})$, the innovation and its covariance are
\begin{equation}
\boldsymbol{\nu}_k
=
\mathbf{z}_k
-
\mathbf{h}(\hat{\boldsymbol{\xi}}_k^-),
\qquad
\mathbf{S}_k
=
\mathbf{H}_k
\mathbf{P}_k^-
\mathbf{H}_k^\top
+
\mathbf{R}_k,
\label{eq:ekf_innovation}
\end{equation}
with angular innovations wrapped to $[-\pi,\pi]$. We accept the update only if the squared
Mahalanobis distance is below a per-sensor threshold (Table~\ref{tab:loc_params}),
\begin{equation}
d_k^2
=
\boldsymbol{\nu}_k^\top
\mathbf{S}_k^{-1}
\boldsymbol{\nu}_k
<
\tau_s^2.
\label{eq:maha_gate}
\end{equation}
The measurement noise $\mathbf{R}_k$ is the sensor-reported covariance when available, including for measured GNSS yaw, and a fixed per-sensor default otherwise.

Heading estimation prefers the dual-antenna GNSS yaw, used whenever the receiver supplies a valid orientation measurement passing the same innovation gate. When direct GNSS orientation is unavailable, we derive a second heading from the GNSS trajectory and inject it as a yaw-only update, while wheel odometry keeps correcting the velocity states. To avoid estimating heading from displacement comparable to the GNSS position noise, a new position enters the heading window only after the robot moves at least $\delta_p$. Up to $N$ such positions are retained (Table~\ref{tab:loc_params}). We circularly average the directions $\beta_j$ of the steps between consecutive accepted positions over the retained GNSS window,
\begin{equation}
\hat{\psi}_k
=
\operatorname{atan2}
\left(
\sum_j \sin\beta_j,
\sum_j \cos\beta_j
\right).
\label{eq:course}
\end{equation}
The fallback is accepted only if the window directions' circular dispersion stays
below $V_{\mathrm{thr}}$. If no heading source is available, the EKF skips the yaw update and
propagates heading with its motion model until a valid one appears.

\begin{table}[t]
    \centering
    \caption{Localization parameters used consistently throughout the reported experiments. Process-noise entries are per-axis variance rates, entering Eq.~\eqref{eq:ekf_cov_prediction} scaled by $\Delta t$; $\tau_s$, $N$, and $V_{\mathrm{thr}}$ are dimensionless; every gate value is a Mahalanobis distance.}
    \label{tab:loc_params}
    \vspace{0.6em}
    \scriptsize
    \setlength{\tabcolsep}{4pt}
    \begin{tabular}{@{}ll@{}}
    \toprule
    \textbf{Parameter} & \textbf{Value} \\
    \midrule
    $\mathbf{Q}_{p}$ (position)               & $(0.05,\,0.05,\,0.05)$ \\
    $\mathbf{Q}_{\mathrm{rpy}}$ (orientation)          & $(0.03,\,0.03,\,0.03)$ \\
    $\mathbf{Q}_{v}$ (linear velocity)        & $(0.025,\,0.025,\,0.04)$ \\
    $\mathbf{Q}_{\omega}$ (angular velocity)  & $(0.01,\,0.01,\,0.02)$ \\
    $\tau_{\mathrm{odom,twist}}$              & $10$ \\
    $\tau_{\mathrm{GNSS}}$ (position, measured yaw) & $5$ \\
    $\delta_p$ (minimum step)                 & \SI{0.3}{\metre} \\
    $N$ (window size)                         & $5$ \\
    $V_{\mathrm{thr}}$ (consistency bound)    & $0.1$ \\
    \bottomrule
    \end{tabular}
\end{table}

\section{Coverage Planning}
\label{sec:planning}

The offline stage produces the reusable coverage route in two parts: it prepares the admissible region from the site survey, then plans swaths over it with Fields2Cover and our controller-aware refinements before online execution.

\subsection{Coverage Preparation}
\label{subsec:perception_mapping}

Preparation yields the admissible region $\mathcal{A}$, placing the boundary and known obstacles in the metric frame used by the coverage planner and navigation stack.

\emph{Perimeter acquisition} runs once per site or layout change by either driving the robot along the boundary while logging georeferenced poses or manually entering known waypoints during site setup. The resulting poses are subsampled by translation and yaw thresholds, optionally edited for consistency, and accepted only if they form a valid, closed, non-self-intersecting polygon $\mathcal{B}$.

\emph{Obstacle preparation} then takes the known-obstacle footprints from a 2D map, a sliced 3D model, or a projected reconstruction and places them metrically with a similarity transformation from pixel-to-world correspondences. It then buffers them by the safety margin $\delta$ (Sect.~\ref{sec:problem}) and subtracts them from $\mathcal{B}$, yielding the polygonal admissible model $\mathcal{A}$ used by all later stages.

\subsection{Controller-Aware Refinements}
\label{subsec:full_coverage}

We build on F2C (Sect.~\ref{sec:related}). Three refinements adapt F2C's geometrically valid output for closed-loop execution, each formalized against the native behavior it replaces. Given $\mathcal{A}$ and the vehicle parameters $(d_s, r_{\min})$, F2C runs four deterministic stages in our setup: it (i)~offsets the boundary inward to reserve a constant-width \emph{headland} for turning; (ii)~fills the interior with parallel \emph{swaths} at spacing $d_s$; (iii)~orders the swaths into a route $\Sigma=(\sigma_1,\sigma_2,\dots)$ minimizing non-productive travel, with orderings such as boustrophedon, snake, and spiral; and (iv)~joins consecutive swaths with motion primitives (Dubins for forward-only motion, Reeds-Shepp when reverse is allowed) into a drivable path $\gamma_{\mathrm{F2C}}$. F2C is configured with $d_s$ rather than $w$, which is how the $w-d_s$ overlap between adjacent passes is realized. F2C is geometrically sound overall, but $\gamma_{\mathrm{F2C}}$ does not target closed-loop trackability or clearance at swath transitions.

\subsubsection{Collinearity-Aware Resampling.} 
\begin{figure}[t]
    \centering
    \begin{subfigure}{0.53\linewidth}\centering
        \includegraphics[width=\linewidth]{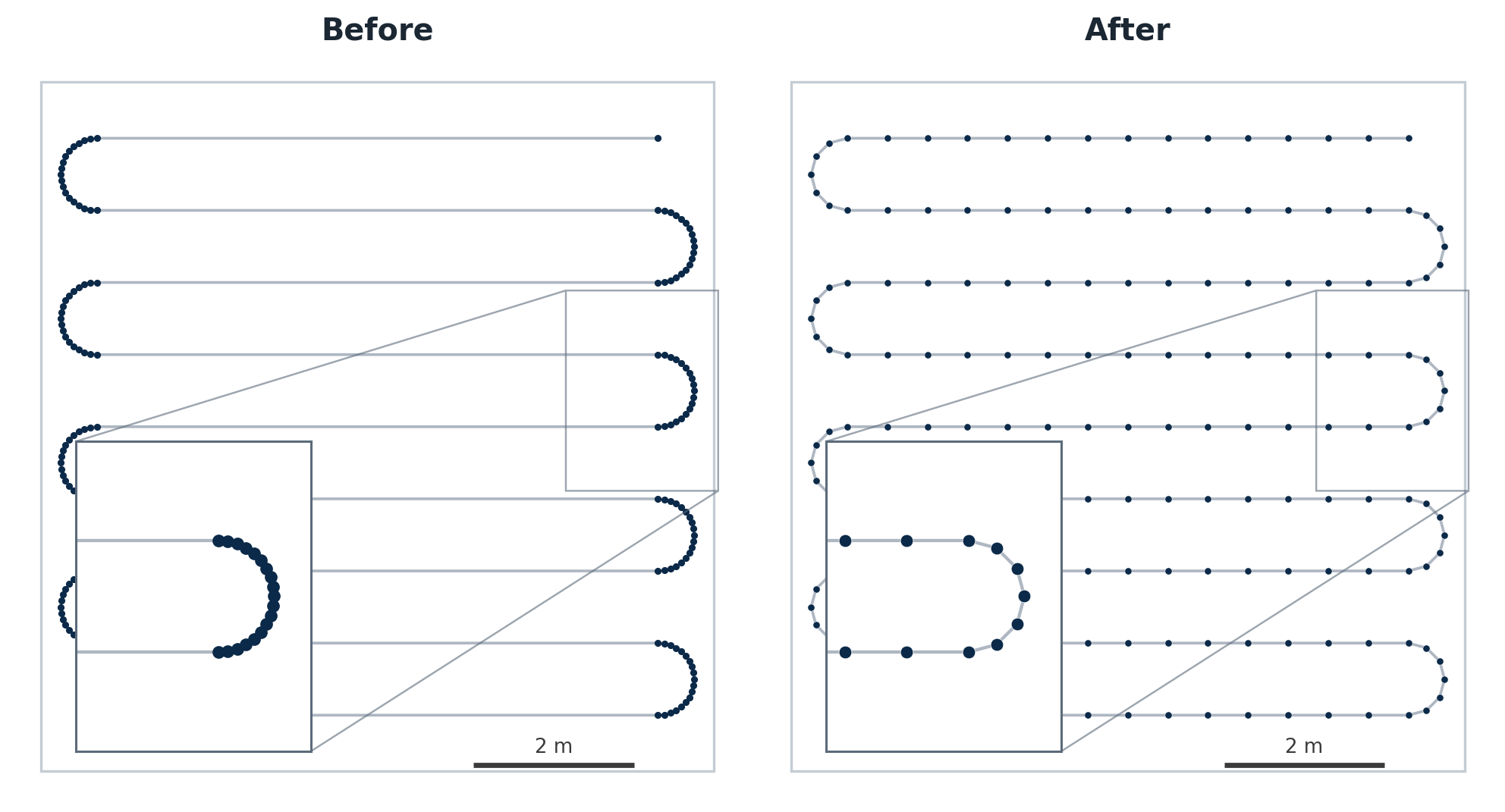}
        \caption{Collinearity-aware resampling.}
        \label{figure:f2c_subsampling}
    \end{subfigure}\hfill
    \begin{subfigure}{0.46\linewidth}\centering
        \includegraphics[width=\linewidth]{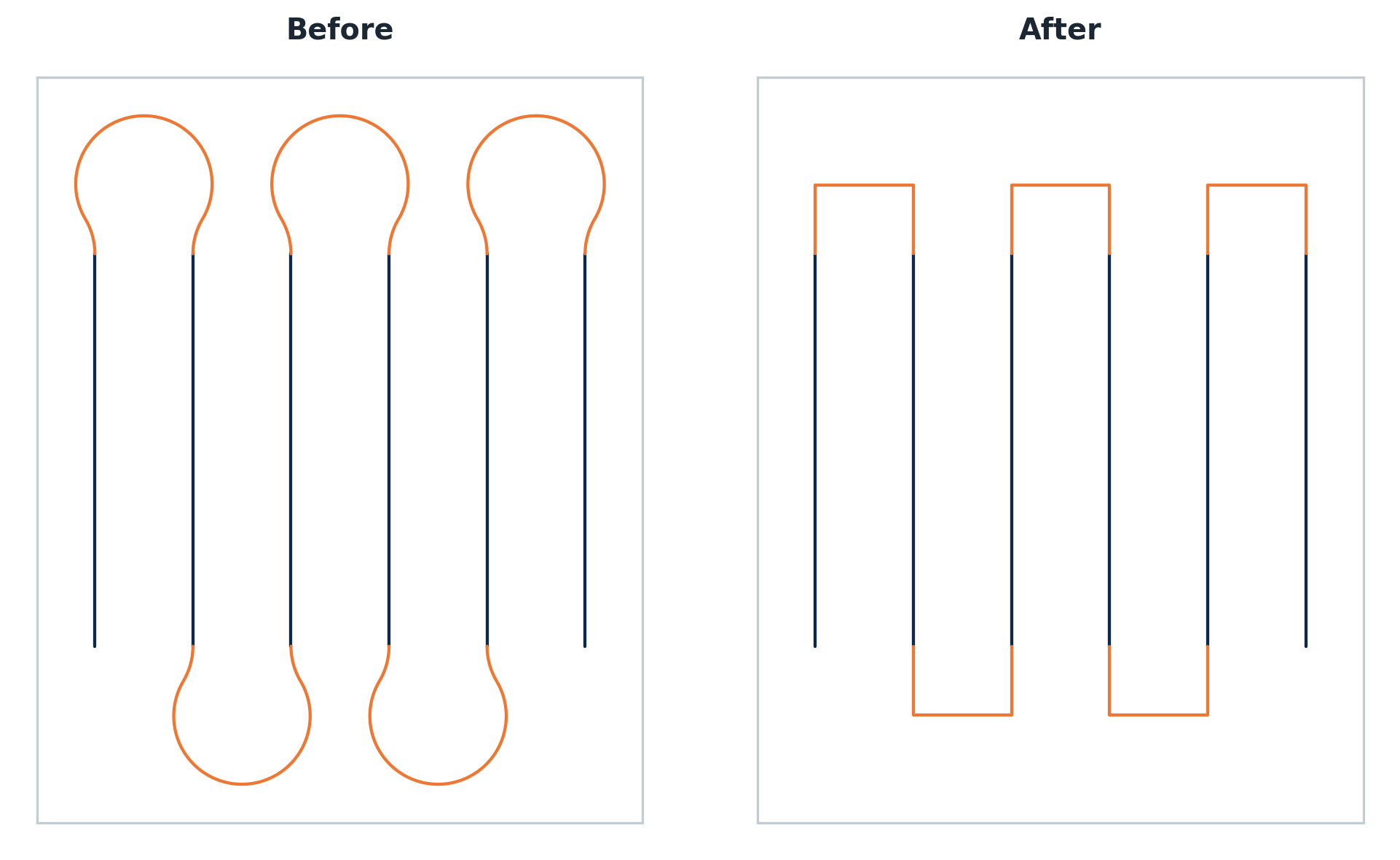}
        \caption{Obstacle-aware connectors.}
        \label{figure:connectors}
    \end{subfigure}
    \caption{Trajectory refinements to F2C, before (left) and after (right): (a)~resampling equalizes waypoint density on straights and keeps turns; (b)~F2C's looping connectors become straight links realized online by the reactive layer at execution time.}
    \label{figure:refinements}
\end{figure}

F2C samples turns densely but can leave long straight swaths with few intermediate poses, degrading tracking and overlap where coverage matters most. Uniform resampling would fix the straights but dilute F2C's dense turn sampling, so we instead adapt waypoint density to the local geometry without altering the path traced. For an ordered waypoint sequence $W=(p_1,\dots,p_n)$ with $p_i\in\mathbb{R}^2$, let $c_i=p_{i+1}-p_{i-1}$ be the chord skipping vertex $p_i$ and $d_i=p_i-p_{i-1}$ the incoming step; the angle between them,
\begin{equation}
  \alpha_i = \arccos\frac{c_i\cdot d_i}{\lVert c_i\rVert\,\lVert d_i\rVert},
  \label{eq:collinear}
\end{equation}
is zero when $p_i$ lies on the chord and grows as the path bends more sharply there. Resampling runs three passes (Algorithm~\ref{alg:resample}, parameters in Table~\ref{tab:exec_params}): a distance pass keeping one point per $d_{\min}$ and splitting gaps above $d_{\max}$; a decimation pass dropping near-collinear vertices ($\alpha_i\le\alpha_{\mathrm{thr}}$); and a final pass re-interpolating gaps still above $d_{\max}$. Turns keep a dense sampling while straight swaths end up uniformly sampled at $d_{\max}$; points are added or dropped only on near-straight stretches or below $d_{\min}$, so resampling leaves the path and its swept ground essentially unchanged, changing only the goal density consumed during execution (Fig.~\ref{figure:f2c_subsampling}), which is what the downstream controller sees.

\subsubsection{Obstacle-Aware Connectors.} In F2C the swaths respect the clearance margin $\delta$ (Sect.~\ref{sec:problem}), but the connectors joining consecutive swaths come from a separate routine that need not do so; in cluttered layouts they can cross an inflated obstacle $O_i^{\delta}$, making the offline plan infeasible (Fig.~\ref{figure:connectors}, left). We therefore disable native connector generation and join each swath $\sigma_k$ to the next by a straight link $\ell_k$ from the end of $\sigma_k$ to the start of $\sigma_{k+1}$ (Fig.~\ref{figure:connectors}, right). This link is a goal-to-goal \emph{intent}, not a committed trajectory: it fixes a deterministic structure between swaths without an offline feasibility check, because at run time the global planner routes around known obstacles in the global costmap, while the controller enforces clearance against unknown ones on live perception. An obstructed link is absorbed by the goal-management policy of Sect.~\ref{subsec:obstacles}, which defers the affected goals and usually covers the region on a later pass, or drops them once the postponement budget is exhausted during the mission.

\subsubsection{Layout-Aware Obstacle Clustering.} Densely packed small obstacles can create many gaps too narrow for the robot to traverse safely, leading F2C to generate fragile transitions. We therefore merge nearby obstacles with a morphological operation: the obstacle layer is rasterized into a binary occupancy image and dilated with a rectangular structuring element, i.e., a small mask that expands occupied pixels. The mask is wider across the driving direction than along it and is sized according to the robot clearance. Outer contours of the dilated occupied regions then yield the merged obstacle polygons. Two obstacles merge when their gap is smaller than the mask width in the connection direction, so one rasterize-dilate-contour pass both enforces the safety margin and removes sub-traversable gaps. The absorbed area is limited to those narrow gaps, so the admissible region shrinks only slightly while the merged layout keeps generated trajectories collision-free and feasible to execute in the field.

\begin{algorithm}[t]
\footnotesize
\caption{Collinearity-aware resampling}
\label{alg:resample}
\begin{algorithmic}[1]
\Require waypoints $W=(p_1,\dots,p_n)$, spacing bounds $d_{\min}<d_{\max}$, angle $\alpha_{\mathrm{thr}}$
\Ensure resampled waypoints $W'$ tracing the same path
\State $W' \gets (p_1)$
\For{$i = 2$ \textbf{to} $n$}
    \State $\ell \gets \lVert p_i - \operatorname{last}(W')\rVert$
    \If{$\ell > d_{\max}$}
        \State append evenly spaced points (gap $\le d_{\max}$) from $\operatorname{last}(W')$ to $p_i$, then $p_i$
    \ElsIf{$\ell \ge d_{\min}$}
        \State append $p_i$ to $W'$
    \EndIf
\EndFor
\State drop every interior vertex $p_i$ of $W'$ with $\alpha_i \le \alpha_{\mathrm{thr}}$
\State re-insert evenly spaced points along every remaining segment longer than $d_{\max}$
\State \Return $W'$
\end{algorithmic}
\end{algorithm}

\begin{table}[t]
    \centering
    \caption{Coverage-planning and execution parameters used in the experiments.}
    \label{tab:exec_params}
    \vspace{0.6em}
    \scriptsize
    \setlength{\tabcolsep}{4pt}
    \begin{tabular}{@{}lll@{}}
    \toprule
    \textbf{Stage} & \textbf{Parameter} & \textbf{Value} \\
    \midrule
    \multirow{5}{*}{Coverage planning}
      & $w$ (working width)                     & \SI{1.0}{\metre} \\
      & $d_s$ (swath spacing)                   & \SI{0.9}{\metre} \\
      & $d_{\min}$ (minimum waypoint spacing)   & \SI{0.10}{\metre} \\
      & $d_{\max}$ (maximum waypoint spacing)   & \SI{0.5}{\metre} \\
      & $\alpha_{\mathrm{thr}}$ (collinearity)  & \SI{0.01}{\degree} \\
    \midrule
    \multirow{6}{*}{Execution}
      & $K$ (goal horizon)                      & $4$ \\
      & goal pruning radius                     & \SI{0.85}{\metre} \\
      & controller path truncation              & \SI{5}{\metre} \\
      & $c_{\max}$ (goal cost threshold)        & $10$ \\
      & $r_{\max}$ (postponements)              & $3$ \\
      & goal tolerance (position, heading)      & \SI{0.5}{\metre}, \SI{0.5}{\radian} \\
    \bottomrule
    \end{tabular}
\end{table}

\section{Online Execution}
\label{subsec:autonomous_navigation}

The online stack targets long-horizon coverage in large outdoor sites, where the robot must follow long waypoint sequences, avoid obstacles safely, and recharge autonomously. In a layered \emph{sense-localize-plan-act} structure over the localization stack of Sect.~\ref{subsec:localization}, a supervising behavior tree coordinates navigation, recovery, and energy management from first undocking to final docking.

\subsection{Behavior-Tree Mission Executive}
\label{subsec:bt}

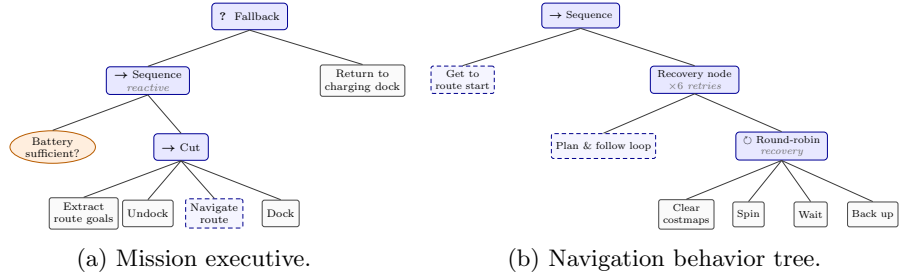
\begin{figure}[t]
\centering
\tikzset{
  ctl/.style ={draw=blue!58!black, fill=blue!9, rounded corners=2.5pt, semithick, align=center, inner xsep=5pt, inner ysep=2.5pt, minimum height=6mm},
  cnd/.style ={draw=orange!72!black, fill=orange!13, ellipse, semithick, align=center, inner xsep=2pt, inner ysep=1pt, minimum height=6mm},
  act/.style ={draw=gray!55!black, fill=gray!4, rounded corners=1.5pt, semithick, align=center, inner sep=3pt, minimum height=6mm},
  sub/.style ={act, draw=blue!55!black, densely dashed, fill=blue!4},
  ed/.style  ={semithick, draw=gray!45!black}}
\newcommand{\btnote}[1]{{\scriptsize\itshape\color{black!58}#1}}
\begin{subfigure}[t]{0.40\linewidth}\centering
\resizebox{!}{3cm}{%
\begin{tikzpicture}[font=\scriptsize, >=Latex]
  \node[ctl] (root) at (0,0)        {\textbf{?}~~Fallback};
  \node[ctl] (pseq) at (-2.30,-1.45){$\boldsymbol{\rightarrow}$~Sequence\\\btnote{reactive}};
  \node[act] (rdk)  at ( 2.55,-1.45){Return to\\charging dock};
  \node[cnd] (batt) at (-4.45,-2.95){Battery\\sufficient?};
  \node[ctl] (cut)  at (-1.55,-2.95){$\boldsymbol{\rightarrow}$~Cut};
  \node[act] (ext)  at (-3.75,-4.45){Extract\\route goals};
  \node[act] (und)  at (-2.30,-4.45){Undock};
  \node[sub] (ntp)  at (-0.80,-4.45){Navigate\\route};
  \node[act] (dk)   at ( 0.70,-4.45){Dock};
  \draw[ed] (root.south)--(pseq.north); \draw[ed] (root.south)--(rdk.north);
  \draw[ed] (pseq.south)--(batt.north); \draw[ed] (pseq.south)--(cut.north);
  \draw[ed] (cut.south)--(ext.north); \draw[ed] (cut.south)--(und.north);
  \draw[ed] (cut.south)--(ntp.north); \draw[ed] (cut.south)--(dk.north);
\end{tikzpicture}}
\caption{Mission executive.}
\label{figure:bt_mission}
\end{subfigure}\hfill
\begin{subfigure}[t]{0.58\linewidth}\centering
\resizebox{!}{3cm}{%
\begin{tikzpicture}[font=\scriptsize, >=Latex]
  \node[ctl] (root) at (0,0)         {$\boldsymbol{\rightarrow}$~Sequence};
  \node[sub] (start)at (-2.85,-1.55) {Get to\\route start};
  \node[ctl] (rec)  at ( 2.75,-1.55) {Recovery node\\\btnote{$\times6$ retries}};
  \node[sub] (pf)   at ( 0.55,-3.15) {Plan \& follow loop};
  \node[ctl] (rr)   at ( 4.85,-3.15) {$\boldsymbol{\circlearrowright}$~Round-robin\\\btnote{recovery}};
  \node[act] (clr)  at ( 2.55,-4.80) {Clear\\costmaps};
  \node[act] (spin) at ( 4.05,-4.80) {Spin};
  \node[act] (wait) at ( 5.55,-4.80) {Wait};
  \node[act] (bk)   at ( 7.05,-4.80) {Back up};
  \draw[ed] (root.south)--(start.north); \draw[ed] (root.south)--(rec.north);
  \draw[ed] (rec.south)--(pf.north);     \draw[ed] (rec.south)--(rr.north);
  \draw[ed] (rr.south)--(clr.north); \draw[ed] (rr.south)--(spin.north);
  \draw[ed] (rr.south)--(wait.north);\draw[ed] (rr.south)--(bk.north);
\end{tikzpicture}}
\caption{Navigation behavior tree.}
\label{figure:bt_nav}
\end{subfigure}
\caption{Two-level behavior-tree execution. (a)~The executive runs the deployment loop while the battery suffices and otherwise returns to the dock. (b)~The Nav2 tree drives to the route start, then wraps the plan-and-follow loop in a recovery node of escalating actions. Ellipses are conditions, solid boxes actions, dashed boxes subtrees.}\label{figure:bt}
\end{figure}

Mission execution uses a two-level behavior tree (Fig.~\ref{figure:bt}). A compact \emph{mission tree} runs the deployment loop: while the battery suffices, it extracts the route goals, undocks, hands the whole route to the navigation layer, and docks again; low battery or a failed mission returns the robot to the dock. The navigation layer is a two-stage Nav2 BT: a \emph{get-to-route} phase driving to the first swath with a single-goal horizon, then a recovery-wrapped \emph{plan-and-follow loop} for the remaining route. The loop plans over a short horizon of the next $K=4$ goals, prunes each once the robot passes within \SI{0.85}{\metre} of it, truncates the controller path to \SI{5}{\metre} ahead, and re-smooths it every control cycle, so replanning stays bounded and local for any route length or goal count. The pruning radius, between $d_{\max}$ and the swath spacing $d_s$ (Table~\ref{tab:exec_params}), keeps motion continuous while preventing the controller from reaching a goal on the adjacent pass.

Robustness comes from a recovery subtree of escalating actions (Fig.~\ref{figure:bt}), tried one at a time under a bounded retry budget: clearing the local then global costmap, in-place rotation to refresh perception, a short wait for unknown obstacles to clear, and a short controlled back-up to free the robot when stuck.

\subsection{Reactive Obstacle Handling and Goal Management}
\label{subsec:obstacles}

The local costmap inflates LiDAR-observed occupancy into safety buffers discouraging passages too narrow to execute reliably, while the global costmap holds only the operational boundary and known obstacles from preparation (Sect.~\ref{subsec:perception_mapping}), so offline planning sets intent and online perception enforces safety.

We treat unknown obstacles during coverage as goal management, not merely replanning: densely sampled coverage trajectories mean an unknown obstacle may overlap upcoming goals, so replanning alone may still drive toward a blocked pose. Before each planning cycle, the supervisor pushes any of the next goals with local-costmap cost above $c_{\max}$ to the back of the queue, so the planner first follows the currently reachable goals. A deferred goal is retried once it returns to the front and dropped only after $r_{\max}$ postponements (Table~\ref{tab:exec_params}). Because the rolling local costmap clears as obstacles disappear, a region blocked on one pass is usually covered on a later one while progress continues safely elsewhere.

\subsection{Autonomous Docking}
\label{subsec:docking}

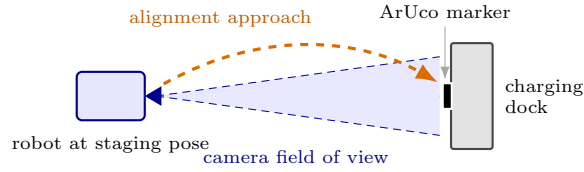
\begin{figure}[t]
\centering
\begin{tikzpicture}[font=\scriptsize, >=Latex]
  \draw[densely dotted, gray!55!black] (1.95,0) -- (5.85,0);
  \fill[blue!9] (2.05,0) -- (5.8,0.52) -- (5.8,-0.52) -- cycle;
  \draw[blue!55!black, densely dashed, very thin] (2.05,0) -- (5.8,0.52);
  \draw[blue!55!black, densely dashed, very thin] (2.05,0) -- (5.8,-0.52);
  \node[anchor=north, text=blue!50!black] at (3.9,-0.60) {camera field of view};
  \draw[thick, draw=gray!55!black, fill=gray!22, rounded corners=1pt] (5.95,-0.7) rectangle (6.5,0.7);
  \node[anchor=west, align=left] at (6.55,0) {charging\\dock};
  \draw[thick, fill=black, draw=white] (5.84,-0.17) rectangle (5.96,0.17);
  \node[anchor=south] (mk) at (5.87,0.92) {ArUco marker};
  \draw[->, gray!70] (mk) -- (5.87,0.22);
  \draw[thick, draw=blue!55!black, fill=blue!10, rounded corners=2pt] (1.0,-0.32) rectangle (1.9,0.32);
  \fill[blue!55!black] (1.9,0) -- (2.12,0.13) -- (2.12,-0.13) -- cycle;
  \node[anchor=north] at (1.45,-0.42) {robot at staging pose};
  \draw[->, very thick, dashed, draw=orange!85!black] (2.15,0.16) .. controls (3.5,0.92) and (4.8,0.74) .. (5.78,0.18);
  \node[anchor=south, text=orange!72!black] at (2.9,0.80) {alignment approach};
\end{tikzpicture}
\caption{Autonomous docking (top view). From a staging pose, the robot follows a perception-guided path (dashed) to an ArUco marker kept in camera view; its power state confirms success. Undocking reverses alignment to the staging pose.}
\label{figure:docking}
\end{figure}

Long-duration operation is supported by autonomous docking integrated into the same BT supervisor, which interrupts coverage for docking once the battery falls below its threshold, so return-to-charge is part of the mission logic rather than a special case outside it. Docking is a two-stage procedure (Fig.~\ref{figure:docking}): the robot first navigates to a predefined \emph{staging pose} at a fixed offset ahead of the charging dock, with a yaw keeping the dock marker visible, then runs a perception-guided alignment phase where a calibrated camera detects the fiducial ArUco marker~\cite{oh2025regression} and estimates its pose as the alignment target. Docking success is inferred from a change in the robot's power state; undocking re-runs the same alignment approach in reverse, backing out to the staging pose.

\section{Evaluation}
\label{sec:eval}

Real-world trials assess the integrated system's mission-level reliability and coverage quality. Of the outdoor tests, we report five representative areas spanning different field shapes and sizes, with and without obstacles.

\subsection{Experimental Setup and Test Scenarios}
\label{subsec:setup}

Evaluation spans both phases: offline preparation (perimeter recording, obstacle extraction, coverage computation; Sects.~\ref{subsec:perception_mapping}, \ref{subsec:full_coverage}) and online execution (route following, multi-goal navigation, recovery, docking, unknown-obstacle avoidance; Sect.~\ref{subsec:autonomous_navigation}). The coverage planner was also run in simulation on a \SI{1.5}{\hectare} photovoltaic site, larger than any test site. Each real-world run exercised the planning-to-execution pipeline under realistic sensing and terrain conditions, autonomously from start to finish, under the same operating protocol and parameter set.

The five areas span up to \SI{1641}{\square\metre}: convex (01, 02, 05) and non-convex (03, 04), with known obstacles (03, 04, 05) and without (01, 02). Known footprints of varying number and size were buffered and subtracted from $\mathcal{B}$ offline; unknown obstacles such as pedestrians or temporarily placed objects were present in some of the trials, with the outcome reported in Sect.~\ref{subsec:results}.

\subsection{Evaluation Metrics}
\label{subsec:metrics}

Besides the coverage efficiency $\eta_c$ of Eq.~\eqref{eq:etac}, \emph{mission success} is a completion criterion, not a per-area quantity: a mission succeeds if, without human intervention after perimeter definition, the robot reaches the route start, traverses the whole route, and completes the terminal phase (docking when required). All five reported missions met this criterion, so Table~\ref{tab:results} reports each run's wall-clock duration instead. Both $\eta_c$ sweeps follow Eq.~\eqref{eq:swept} at working width $w$, clipped to the admissible region $\mathcal{A}$; ground swept twice counts once.

\subsection{Field Results}
\label{subsec:results}

\begin{table}[t]
    \centering
    \caption{Coverage results for the five areas. \emph{Field} is the area enclosed by $\mathcal{B}$, \emph{Planned} the area a perfect run sweeps at working width $w$, \emph{Route} the planned-route length, \emph{Obstacles} the mapped-obstacle area inside the boundary, \emph{Duration} the time from first to last motion, $\eta_c$ the fraction of \emph{Planned} area swept during execution (Eq.~\eqref{eq:etac}).}
    \label{tab:results}
    \vspace{0.6em}
    \small
    \setlength{\tabcolsep}{5pt}
    \resizebox{\textwidth}{!}{%
    \begin{tabular}{clrrrrrr}
    \toprule
    \textbf{Seq.} & \textbf{Shape} & \textbf{Field} & \textbf{Planned} & \textbf{Route} & \textbf{Obstacles} & \textbf{Duration} & $\boldsymbol{\eta_c}$ \\
     &  & [\si{\square\metre}] & [\si{\square\metre}] & [\si{\metre}] & [\si{\square\metre}] & [min:s] & [\si{\percent}] \\
    \midrule
    01 & convex     & 168.0  & 156.3  & 180.0  & 0.0  & 8:35  & 93.2 \\
    02 & convex     & 196.1  & 190.6  & 221.2  & 0.0  & 10:12 & 94.2 \\
    03 & non-convex & 372.7  & 304.2  & 356.3  & 7.5  & 16:07 & 93.1 \\
    04 & non-convex & 948.2  & 711.9  & 837.2  & 65.2 & 36:15 & 94.9 \\
    05 & convex     & 1640.9 & 1397.9 & 1614.1 & 98.5 & 68:37 & 96.1 \\
    \bottomrule
    \end{tabular}%
    }
\end{table}

\begin{figure}[t]
    \centering
    \begin{subfigure}[b]{0.5\linewidth}
      \centering
      \includegraphics[height=3.7cm]{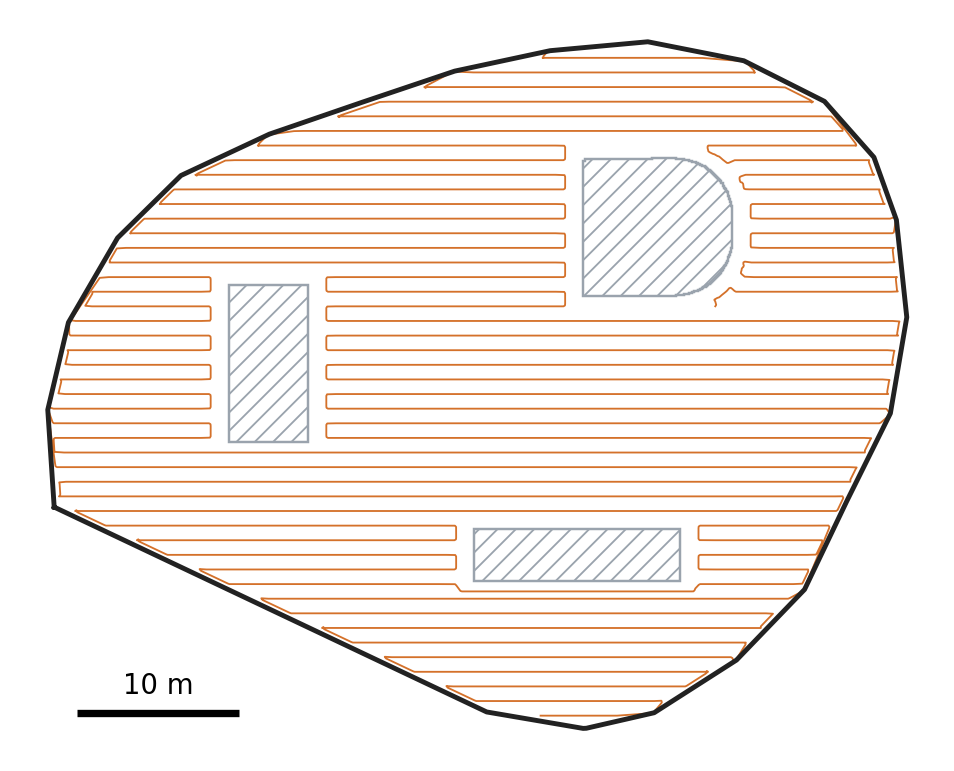}
      \label{fig:res-a}
    \end{subfigure}\hfill
    \begin{subfigure}[b]{0.5\linewidth}
      \centering
      \includegraphics[height=3.7cm]{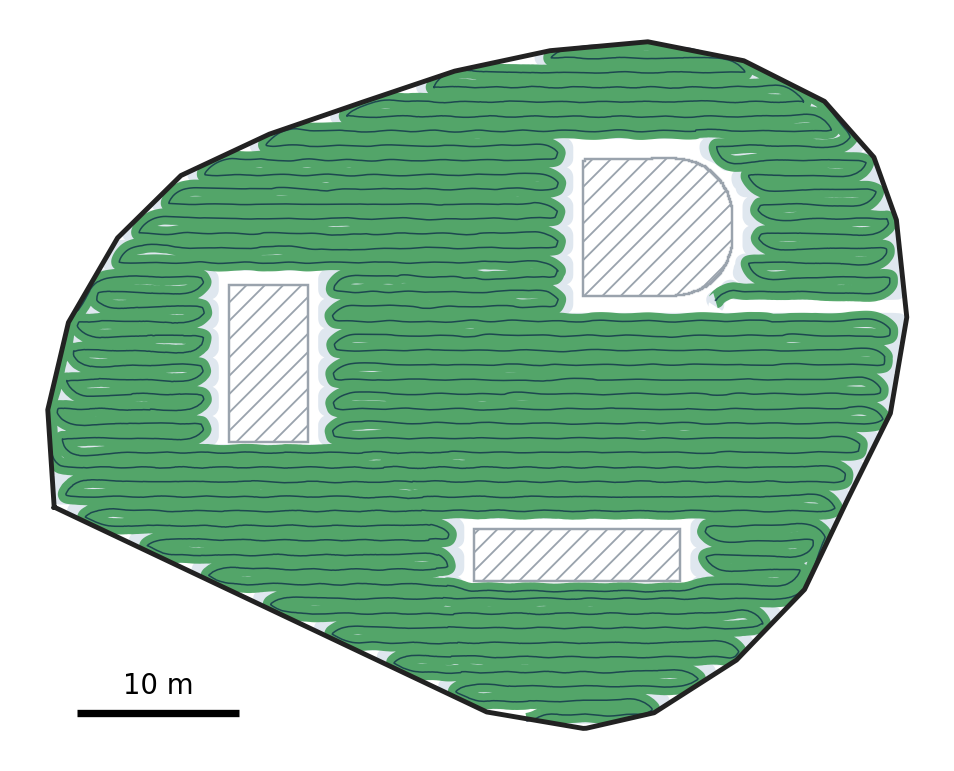}
      \label{fig:res-b}
    \end{subfigure}
    \caption{Planned route versus executed coverage for field Seq.~05 (convex, three obstacles), at matching scale. Left: parallel swaths joined by straight links. Right: executed run, with swept area in green and mapped obstacles shown as striped footprints.}
    \label{fig:coverage_results}
\end{figure}

Fig.~\ref{fig:coverage_results} shows planned swaths followed consistently across a whole representative field (Seq.~05). Across the five areas, the robot successfully completed the full coverage route autonomously in unattended runs of under nine minutes to over an hour, sweeping \SIrange[quantity-product={}]{93.1}{96.1}{\percent} of the planned coverage area. The \emph{planned} passes tile the admissible coverage region without nominal inter-swath gaps: the designed \SI{0.1}{\metre} overlap absorbs small lateral tracking errors in execution, though larger excursions can still leave thin uncovered strips.

Across these five trials, coverage efficiency varied little with scale and clutter. Planned routes grow from \SI{180}{\metre} on the smallest, convex field (Seq.~01) to over \SI{1.6}{\kilo\metre} on the largest, with the most obstacle area (Seq.~05, \SI{1641}{\square\metre}, \SI{98.5}{\square\metre} in three obstacles), yet $\eta_c$ spans only three points across the five areas (Table~\ref{tab:results}), and the non-convex fields with obstacles (Seq.~03--04) track as tightly as the obstacle-free convex ones. Obstacle-aware connectors and layout-aware clustering (Sect.~\ref{subsec:full_coverage}) keep planned routes feasible around the known obstacles recorded at preparation time, and the robot followed those routes reliably in the field.

The route-length column shows a further regularity: in all five areas the planned route runs \num{1.15} to \num{1.18} meters of path per square meter of planned coverage, essentially constant with field size and shape across the test sites. Most of that excess is the designed overlap, which alone costs $1/d_s=\num{1.11}$ meters of path per square meter, so the headland turns and the inter-swath links, taken together, account for only a further \SIrange[quantity-product={}]{4}{6}{\percent} of the planned route.

We tested unknown obstacles directly in the field: pedestrians walked alongside and in front of the robot, and the local costmap, together with the goal-management layer, detected them and maintained clearance. The robot slowed or deferred the blocked goals, routed around them where possible, and resumed coverage once the way was clear, all without manual intervention.

\section{Conclusion and Future Work}

We presented a unified ROS~2 system for single-robot outdoor coverage, combining dual-antenna RTK-GNSS localization, controller-oriented refinements to Fields2Cover, and a behavior-tree executive on Nav2 for long-duration execution with obstacle handling, recovery, and autonomous docking. Across five outdoor areas of different shape, size, and obstacle layout, the system completed every coverage route, the executed sweep tracking the plan closely, while the executive handled unknown obstacles and returned to charge without intervention. Future work will extend it to multi-robot coordination and to localization that stays reliable when GNSS is highly degraded or unavailable for long stretches.

\end{document}